\documentclass[conference]{IEEEtran}
\IEEEoverridecommandlockouts
\usepackage{cite}
\usepackage{amsmath,amssymb,amsfonts}
\usepackage{algorithmic}
\usepackage{graphicx}
\usepackage{textcomp}
\usepackage{xcolor}
\def\BibTeX{{\rm B\kern-.05em{\sc i\kern-.025em b}\kern-.08em
    T\kern-.1667em\lower.7ex\hbox{E}\kern-.125emX}}
\begin{document}

\title{Estimating Body Volume and Height Using 3D Data\\
}

\author{\IEEEauthorblockN{1\textsuperscript{st} Vivek Ganesh Sonar}
\IEEEauthorblockA{\textit{Col. of Engineering \& Computer Science} \\
\textit{Florida Atlantic University}\\
Boca Raton, USA \\
vsonar2023@fau.edu}
\and
\IEEEauthorblockN{2\textsuperscript{nd} Muhammad Tanveer Jan}
\IEEEauthorblockA{\textit{Col. of Engineering \& Computer Science} \\
\textit{Florida Atlantic University}\\
Boca Raton, USA \\
mjan2021@fau.edu}
\and
\IEEEauthorblockN{3\textsuperscript{rd} Mike Wells}
\IEEEauthorblockA{\textit{Charles E. Schmidt Col. of Medicine} \\
\textit{Florida Atlantic University}\\
Boca Raton, USA \\
wellsm@health.fau.edu}
\and
\IEEEauthorblockN{4\textsuperscript{th} Abhijit Pandya}
\IEEEauthorblockA{\textit{Col. of Engineering \& Computer Science} \\
\textit{Florida Atlantic University}\\
Boca Raton, USA \\
pandya@fau.edu}
\and
\IEEEauthorblockN{5\textsuperscript{th} Gabriella Engstrom}
\IEEEauthorblockA{\textit{Charles E. Schmidt Col. of Medicine} \\
\textit{Florida Atlantic University}\\
Boca Raton, USA \\
gengstr1@health.fau.edu}
\and
\IEEEauthorblockN{6\textsuperscript{th} Richard Shih}
\IEEEauthorblockA{\textit{Charles E. Schmidt Col. of Medicine} \\
\textit{Florida Atlantic University}\\
Boca Raton, USA \\
rshih@health.fau.edu}
\and
\IEEEauthorblockN{7\textsuperscript{th} Borko Furht}
\IEEEauthorblockA{\textit{Col. of Engineering \& Computer Science} \\
\textit{Florida Atlantic University}\\
Boca Raton, USA \\
bfurht@fau.edu}
}

\maketitle

\begin{abstract}
Accurate body weight estimation is critical in emergency medicine for proper dosing of weight-based medications, yet direct measurement is often impractical in urgent situations. This paper presents a non-invasive method for estimating body weight by calculating total body volume and height using 3D imaging technology. A RealSense D415 camera is employed to capture high-resolution depth maps of the patient, from which 3D models are generated. The Convex Hull Algorithm is then applied to calculate the total body volume, with enhanced accuracy achieved by segmenting the point cloud data into multiple sections and summing their individual volumes. The height is derived from the 3D model by identifying the distance between key points on the body. This combined approach provides an accurate estimate of body weight, improving the reliability of medical interventions where precise weight data is unavailable. The proposed method demonstrates significant potential to enhance patient safety and treatment outcomes in emergency settings.
\end{abstract}



\section{Introduction}
A precise assessment of one's height and body weight is crucial in several fields such as healthcare, exercise, and ergonomics. Existing conventional approaches depend on direct measuring equipment, which can be laborious and necessitate direct physical interaction with the individual being monitored. Technological advancements in three-dimensional imaging and machine learning have enabled the creation of non-invasive automated systems that can accurately compute body dimensions. The objective of this work is to explore the feasibility of approximating human weight and height employing three-dimensional data acquired from advanced sensors like the RealSense D415 camera. The integration of computer vision techniques with deep learning models enables the extraction of precise body measurements from 3D data, providing a system that is exceptionally efficient and scalable.

\section{Literature Review}
In emergency care, accurate weight estimation is essential for proper medication dosing, but direct measurement is often impractical in such critical situations. Recent advancements in 3D camera systems, particularly those driven by artificial intelligence, offer a promising solution to this challenge. A systematic review of 14 studies published between 2012 and 2024, primarily involving the use of Microsoft Kinect cameras, demonstrated that these systems could achieve a high level of accuracy in weight estimation. Specifically, many of these systems had 90\% of their estimates within 10\% of the actual weight, indicating significant potential for improving drug dosing accuracy and patient safety in emergency settings. These findings underscore the relevance of 3D camera technology in clinical practice, suggesting that with further validation in larger studies, these systems could become a standard tool for weight estimation in urgent care environments\cite{b01}.

Recent advancements in 3D body scanning technology and machine learning have shown promising results in estimating body weight and height. Jan et al \cite{b02} have compared several studies for body weight estimation using anthropemetric, 2D, 3D and fusion of multiple sensors in a well manner to brigde the gap between all the research studies. Machine learning (ML) algorithms have increasingly been employed in weight estimation tasks due to their capability to learn complex patterns directly from data. Techniques such as linear regression, support vector regression, and neural network-based models are commonly utilized to predict weight from extracted image features. For example, Rativa et al.\cite{b1} explored the use of ML regression techniques to estimate height and weight from anthropometric measurements, aiming to improve the precision and reliability of these estimates for applications in healthcare, fitness, and biometric identification. Similarly, Khan et al. \cite{b2} focused on infant birth weight estimation and low birth weight classification in the United Arab Emirates using ML algorithms. This study holds promise for enhancing prenatal care by enabling early interventions for infants at risk of low birth weight, thereby contributing to improved maternal and child health outcomes.

In addition to these developments, recent studies have explored the integration of traditional weight estimation methods with computer vision techniques to enhance accuracy and efficiency. Fitriyah et al.\cite{b3} developed a weight estimation system using a 2D snapshot of a person’s front pose, applying multiple linear regressions based on body features such as height, shoulder width, abdomen and arm width, and feet width. The study demonstrated the necessity of incorporating all available body features to achieve more accurate weight estimation. In a related study, Dantcheva et al.\cite{b4} investigated the use of 2D facial snapshots to estimate a person’s weight, height, and BMI, employing ResNet-50 architecture for the regression process. The findings suggested that facial images could provide valuable information for these estimates, although weight estimation accuracy was challenged by factors such as weight fluctuations, cosmetic alterations, and image modifications.

Advancements in deep learning, particularly in convolutional neural networks (CNNs), have also led to significant improvements in weight estimation accuracy. Kim \cite{b5} evaluated various weight estimation techniques using force-sensing resistor sensors and deep learning models, finding that the serialization method combined with a deep neural network was the most effective, achieving a mean absolute error of ±4.6 kg. However, challenges such as sensor dead zones and recognition errors were noted, highlighting the need for further refinement. In another study, researchers utilized anthropometric parameters derived from 2D frontal body images to estimate weight, with results indicating that the method was generally effective, although some issues arose with BMI estimation for certain age groups.

Efforts to incorporate multi-modal fusion techniques, combining data from various sources such as depth sensors, thermal imaging, and 3D body scans, have been made to improve the robustness and generalization of weight estimation systems. For instance, Dane et al.\cite{b6} demonstrated the use of a 3D camera and infrared imaging, combined with machine learning algorithms, to estimate patient height and weight with high accuracy, achieving a 2.0\% error in height and a 5.1\% error in weight estimations. The study noted that the 3D camera tended to overestimate weight in overweight patients and underestimate it in underweight patients, though overall accuracy remained high. Another study by Cook et al. \cite{b7}tested the use of a Microsoft Kinect RGB camera with an infrared depth sensor to estimate patient volume, which correlated well with actual weight, despite challenges such as arm positioning that affected the accuracy of volume and density estimates.

Emerging techniques for body weight and height estimation continue to evolve, with innovations such as contactless weight estimation and individualized computed tomography (CT) offering new possibilities. Labati et al.\cite{b8} presented a contactless, low-cost weight estimation method using frame sequences of a walking individual, showing potential for view-independent results without the need for complex body modeling. Furthermore, Geissler et al.\cite{b9} introduced an advanced CT-based technique relying on a simulated digital twin, combined with AI algorithms, to automatically calculate both height and weight. This approach offers a more precise method for individualized medical evaluations, though further research is required to assess its performance across different clinical settings and patient populations.

The integration of machine learning, traditional techniques, deep learning architectures, and multi-modal fusion has led to significant advancements in the field of body weight and height estimation. While promising results have been achieved, ongoing challenges such as accuracy across diverse populations, the influence of body posture and clothing, and the need for more robust datasets indicate areas for future research. Continued exploration of emerging technologies and refinement of existing methods will be crucial for developing more accurate and reliable body metric estimation systems.

\section{Methodology}
The methodology for estimating body weight and height involves several key steps:
\subsection{Data Acquisition}
In order to obtain precise 3D photos, the RealSense D415 camera is placed exactly above the subject, as depicted in Figure 1. This deliberate positioning guarantees that the camera has an unambiguous and unhindered perspective of the whole body, enabling it to precisely record all essential details. The D415 is renowned for its capacity to generate high-resolution depth maps, which constitute intricate depictions of the spatial separation between the camera and different locations on the subject's body. The depth maps offer a three-dimensional perspective of the subject, emphasizing the many contours and forms of the body. The importance of the high resolution of the depth map lies in its ability to capture intricate details, such as subtle curves or edges, which are crucial for accurate measurements. After capturing these 3D images, they are subjected to additional processing, including cleaning, refining, and preparing the data for subsequent stages of our research, such as segmentation and volume quantification.

This processing guarantees the utmost accuracy and usability of the depth information, therefore establishing a strong basis for the subsequent body measuring applications. Using the RealSense D415 camera, we can acquire precise and dependable 3D data, crucial for obtaining unambiguous estimations of body weight and height.

\begin{figure*}
    \centering
    \includegraphics[width=0.25\textwidth]{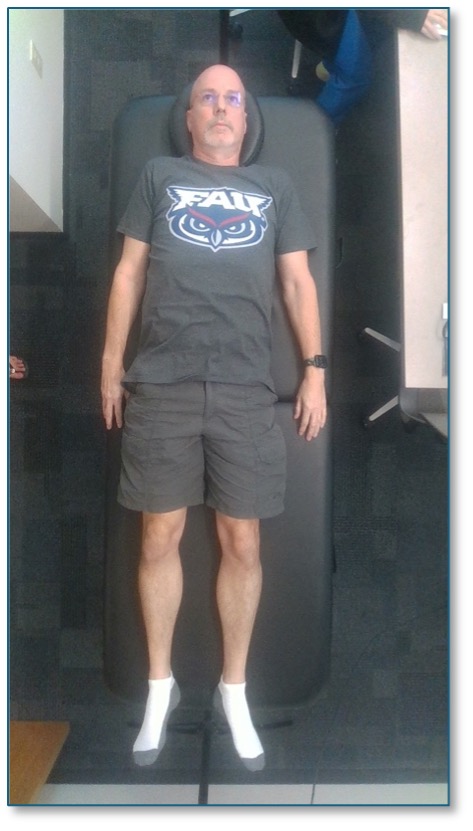}\includegraphics[width=0.25\textwidth]{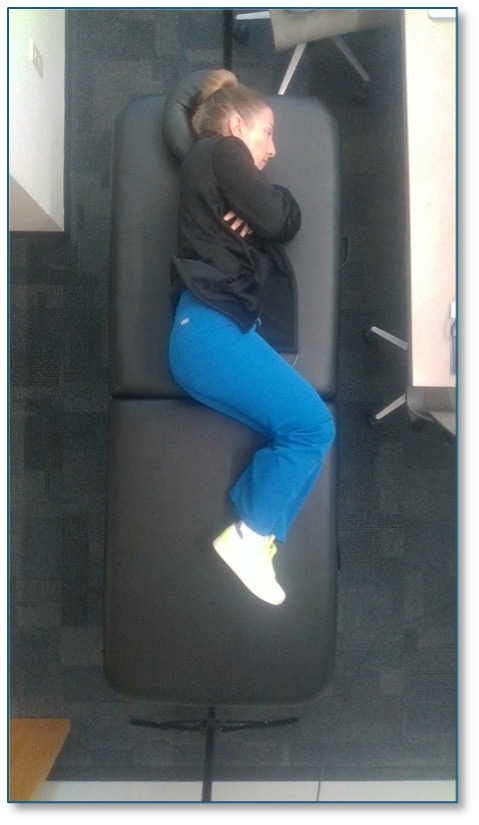}\includegraphics[width=0.25\textwidth]{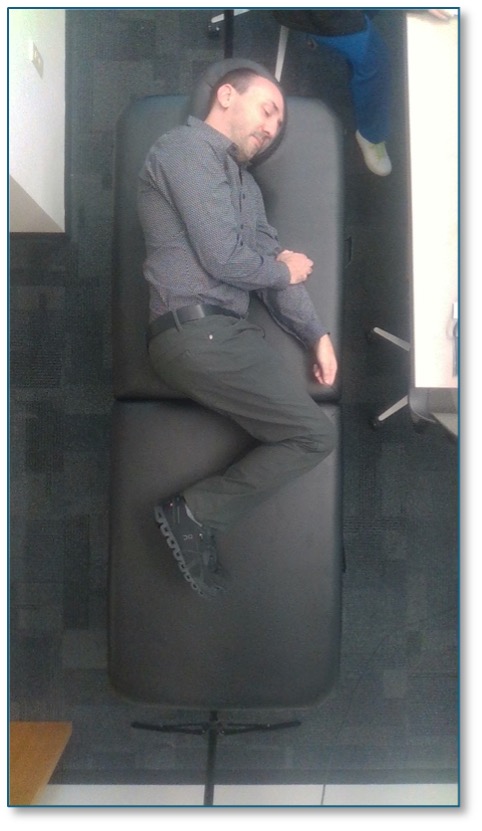}\includegraphics[width=0.25\textwidth]{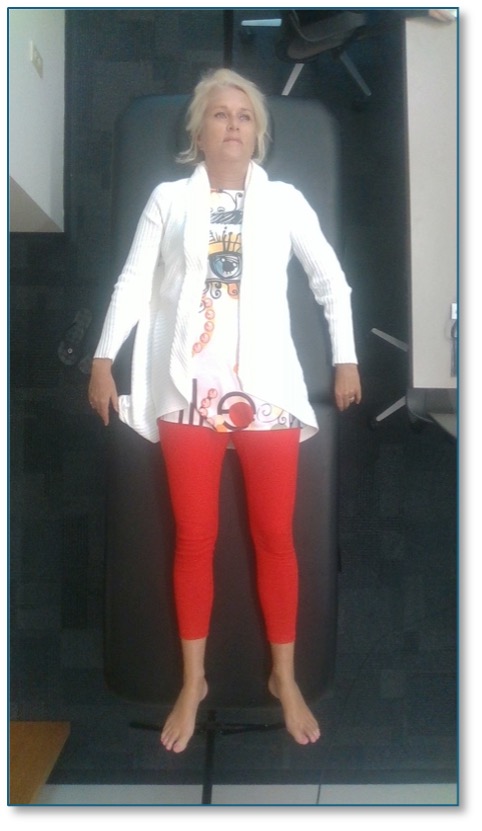}
    \caption{Data acquisition of participants}
    \label{fig:7}
\end{figure*}
\subsection{Image Segmentation}
By utilizing the DepthQualityTool SDK, we initially apply a processing technique to the images obtained by the 3D camera in order to eliminate superfluous components such as the bed or background, therefore allowing the subject to be the sole focal point. The significance of this stage lies in its ability to effectively cleanse the data, therefore facilitating its operational manipulation. Furthermore, we proceed to segment the 3D data, whereby we dissect the image into distinct body components such as the head, arms, and legs.To accomplish this, we employ models such as MobileNet and ResNet, which are neural network architectures specifically developed to identify and segregate distinct components of a picture. The MobileNet model is highly effective at rapidly and effectively detecting anatomical structures, particularly in situations when computational resources are limited. ResNet, however, is more robust and capable of managing more intricate jobs, therefore ensuring precise identification of bodily parts.

Through meticulous processing and segmentation of the data using these methods, we ensure that our measurements of body volume or height are based on clean and well defined data, so enhancing the overall precision of our results.
\begin{figure*}
    \centering
    \includegraphics[width=0.33\textwidth]{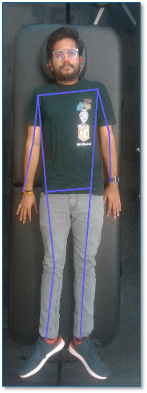}\includegraphics[width=0.33\textwidth]{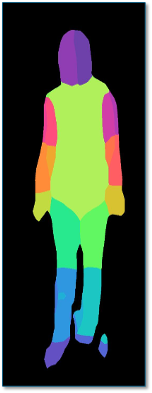}\includegraphics[width=0.33\textwidth]{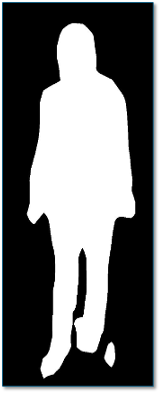}
    \caption{Body pose estimation and body parts segmentation}
    \label{fig:8}
\end{figure*}
\subsection{3D Model Construction}
The 2D photos obtained from the camera are transformed into a 3D model by utilizing a polygon file, which functions as a geometric depiction of the subject's body in three dimensions. This method entails the transformation of 2D pixel data into a 3D coordinate system, enabling the reconstruction of the subject's form and structure. During this conversion, substantial interference, such as spurious data points, discrepancies in depth measurements, and aberrations from the imaging procedure, can be introduced, therefore distorting the precision of the 3D model. Therefore, we employ sophisticated noise reduction methods, such as filtering and smoothing algorithms, to methodically detect and eliminate these undesired components. In this manner, the refinement of the 3D model guarantees that the next procedures, such as volume estimates and feature extraction, rely on accurate and pristine data. Consequently, the overall precision and dependability of the weight and height estimation process are improved.

\subsection{Volume Estimation}
The body volume is calculated using the Convex Hull Algorithm, a technique employed to generate the smallest feasible 3D form, known as a "bounding box," capable of encompassing all the points that represent the body included in the point cloud data. A point cloud is a structured set of data points that precisely delineate the three-dimensional surface of a body. In order to enhance the precision of our volume simulations, we do not employ the Convex Hull Algorithm on the whole body simultaneously. Instead, we partition the point cloud data into several smaller parts measured along the length of the body. Individual segments correspond to distinct anatomical regions of the body, such as the cranium, central body, or extremities. By employing this method, we can compute the volume of each segment separately, therefore enabling more accurate measurements. Following the computation of the volume for each segment, we proceed to aggregate all these volumes in order to obtain the overall body volume.

The advantage of this segmented technique lies in its ability to more precisely factor in changes in body shape and structure compared to the single calculation of volume. By prioritizing smaller anatomical regions, we mitigate the potential for inaccuracies that may arise from considering the body as a unified and flat entity. Employing this approach guarantees the utmost precision in our ultimate volume computation, a critical factor for the following stages of determining body weight and height.

\begin{figure}
    \centering
    \includegraphics[width=0.5\textwidth]{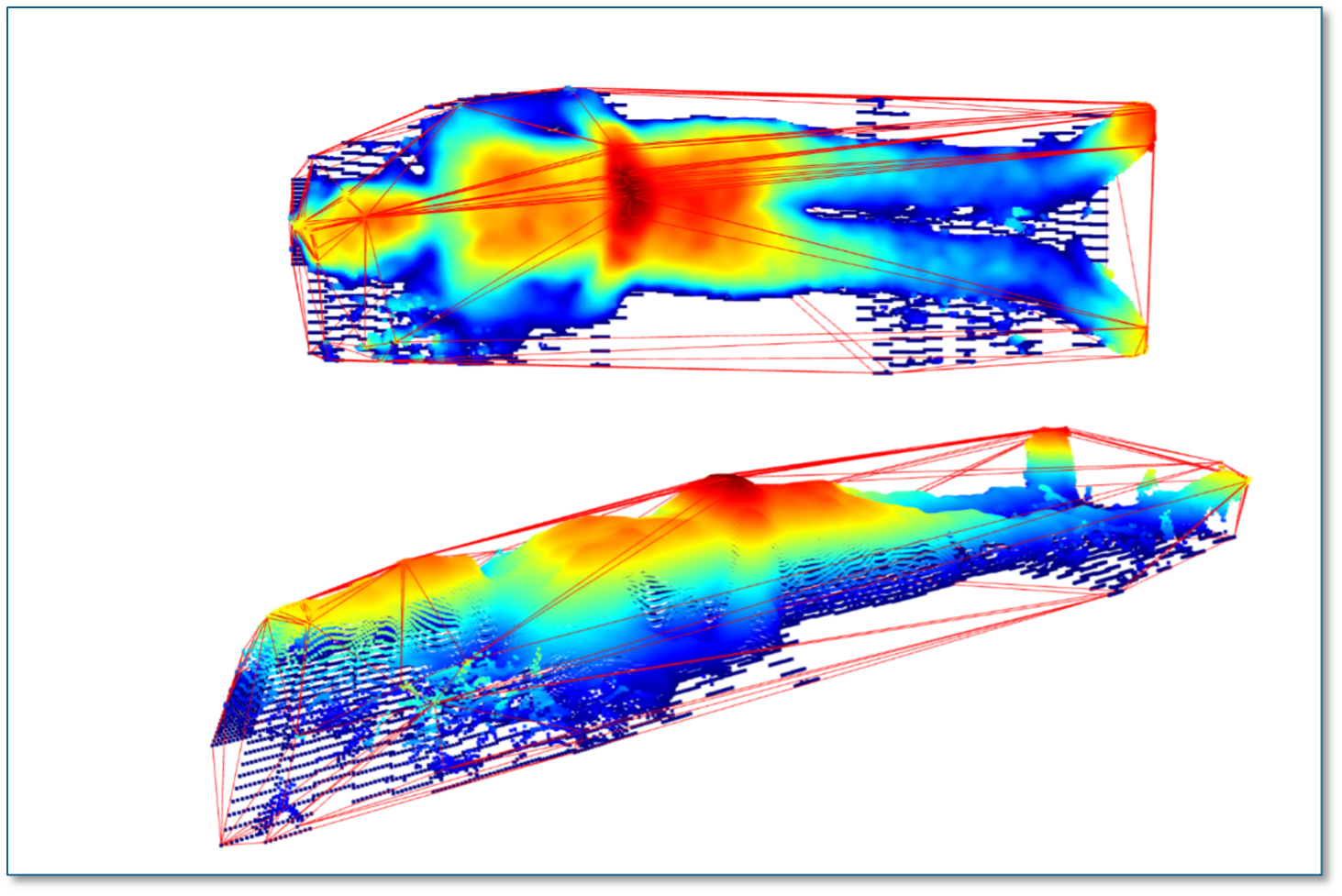}
    \caption{3D image processing using Convex Hull Algorithm}
    \label{fig:5}
\end{figure}

\subsection{Height Estimation}
To determine the height of the subject, the difference between the maximum and minimum Y coordinates within the bounding box that encloses the detected object is computed. The Y coordinates represent the vertical locations of the subject's head and feet, correspondingly. In order to guarantee precision, the height measurement is subsequently modified to account for depth proximity, considering the distance of the subject from the camera. This adjustment of depth serves to offset any distortions in perspective, therefore guaranteeing that the estimated height remains constant and precise irrespective of the subject's position in relation to the camera. To enhance the accuracy of the measurement, the system can incorporate depth data, therefore enabling a more dependable estimation of the actual height of the subject, even in dynamic settings.

\begin{figure}
    \centering
    \includegraphics[width=0.5\textwidth]{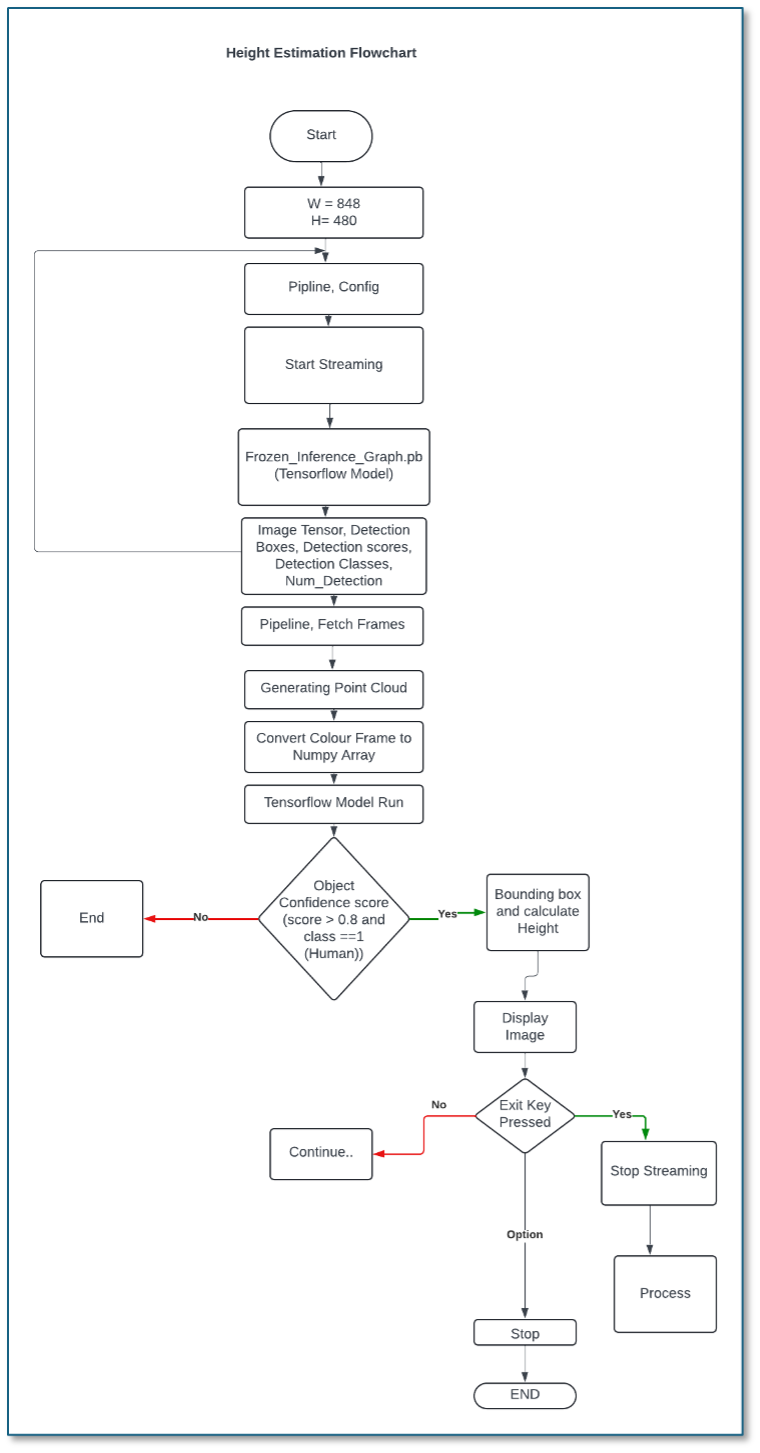}
    \caption{Workflow for estimating patient height}
    \label{fig:workflow}
\end{figure}

\section{Experiments}
The proposed methodology was tested on a dataset of images captured from multiple subjects in different postures. The experiments involved capturing images, segmenting them, and calculating the body volume and height using the described methods. The results were then compared with actual measurements to evaluate the accuracy of the system.

\subsection{Experimental Setup}
In the development of our weight and height estimate system utilising 3D data, we utilised several specialised libraries and tools, carefully selected for their distinct skills in managing intricate computational problems, image processing, and machine learning. Presented below is a detailed explanation of the essential libraries and neural network models employed

\begin{itemize}
    \item Intel RealSense D415: The RealSense D415 camera was used to capture depth data at a resolution of 848x480 pixels, streaming at 30 FPS. The data was processed using TensorFlow models loaded in a pre-trained state. 
    \item NumPy: Designed for numerical computing, Numpy is a Python package that facilitates the manipulation of large, multi-dimensional arrays and matrices using mathematical functions. In our project, Numpy is essential for the processing of 3D data and the execution of matrix multiplication, array manipulation, and other analytical operations. Its exceptional effectiveness and user-friendliness make it indispensable for preprocessing and organizing large datasets for constructing 3D models.
    \item OpenCV: OpenCV is an extensively optimized framework designed for the purpose of real-time computer vision and image processing. Our system utilizes OpenCV to process raw 3D camera images, extract pertinent features, and carry out filtering, edge detection, and contour localization for image segmentation. OpenCV's extensive functionality and wide range of capabilities render it ideal for image transformations and analyses required for accurate segmentation and identification of 3D body parts.
    \item Pillow: Pillow,a Python Imaging Library (PIL), is capable of opening, manipulating, and saving several image file types. The present system effectively handles many photo formats and executes operations such as resizing, format conversion, and fundamental image enhancement. The simplicity and interoperability of Pillow with other Python libraries render it a handy tool for the formatting and preparation of images for processing.
    \item Pyfakewebcam: Pyfakewebcam is a specialized library that enables the development of a virtual webcam. This feature is especially beneficial for the real-time testing and display of segmented images, as it can replicate the output of a physical camera. In our project, Pyfakewebcam enables us to monitor the accuracy of body part identification and alter the parameters accordingly without the need for continuous physical camera input, thereby facilitating the real-time visualization of the segmentation process.
    \item Tensorflow: TensorFlow is a powerful open-source toolkit for deep learning model development and deployment. TensorFlow Estimator simplifies model training and assessment with an API. Our solution builds, trains, and deploys neural networks for image segmentation and other machine learning applications using TensorFlow. Its scalability and flexibility make it perfect for complicated 3D data processing and machine learning algorithm integration.
\end{itemize}

\subsection{Results}
The system showed an average error margin of ±2 cm in height estimation and ±3 kg in weight estimation, demonstrating high accuracy compared to traditional methods. The segmentation and volume estimation processes were key to achieving these results, with the Convex Hull Algorithm proving effective in calculating body volume.

\section{Future Work}
Future research will aim to improve body weight estimation by incorporating more sophisticated models and larger, diverse datasets. This includes integrating different body types to calculate lean body weight (LBW) in adults, accounting for variations in muscle mass, fat distribution, and skeletal structure. Additionally, a refined regression model will estimate weight based on variables like segmented body part volumes, height, age, and gender for more accurate results across diverse populations. The system will also be enhanced for real-time use, with the potential to integrate multiple cameras for better 3D representation and robustness. These advancements will not only improve weight estimation accuracy but also enable the estimation of other metrics such as body fat percentage and muscle mass, making the system more useful in emergency medicine and clinical settings.

\section{Conclusion}
This research demonstrates the potential of using 3D data for accurate estimation of body weight and height. The combination of advanced imaging techniques, noise reduction methods, and deep learning models allows for precise measurement without the need for direct physical contact. As technology advances, these methods could become standard in various fields, offering a more efficient and scalable solution for body metric estimation.

\end{document}